\documentclass{article}
\usepackage{graphicx} % Required for inserting images
\usepackage{amssymb}
\usepackage{amsmath}
\usepackage{float}
\usepackage{svg}
\usepackage{graphicx}
\usepackage{caption}
\usepackage{subcaption}
\usepackage{algorithm}
\usepackage{algpseudocode}
\usepackage[a4paper,margin=2cm]{geometry}

\usepackage{hyperref}

\def\vec#1{\mathchoice{\mbox{\boldmath$\displaystyle#1$}}
{\mbox{\boldmath$\textstyle#1$}} {\mbox{\boldmath$\scriptstyle#1$}} {\mbox{\boldmath$\scriptscriptstyle#1$}}}

\title{Topic Modelling Black Box Optimization}
\author{
Roman Akramov$^{1}$, Artem Khamatullin$^{1}$, Svetlana Glazyrina$^{1}$,\\
Maksim Kryzhanovskiy$^{1,2}$, Roman Ischenko$^{1,2}$\\[0.4em]
\small
$^{1}$Lomonosov Moscow State University\\
\small
$^{2}$Institute for Artificial Intelligence, Lomonosov Moscow State University
\normalsize
}

\date{November 2025}

\begin{document}

\maketitle

\begin{abstract}
Choosing the number of topics $T$ in Latent Dirichlet Allocation (LDA) is a key design decision that strongly affects both the statistical fit and interpretability of topic models. In this work, we formulate the selection of $T$ as a discrete black-box optimization problem, where each function evaluation corresponds to training an LDA model and measuring its validation perplexity. Under a fixed evaluation budget, we compare four families of optimizers: two hand-designed evolutionary methods -- Genetic Algorithm (GA) and Evolution Strategy (ES) -- and two learned, amortized approaches, Preferential Amortized Black-Box Optimization (PABBO) and Sharpness-Aware Black-Box Optimization (SABBO). Our experiments show that, while GA, ES, PABBO, and SABBO eventually reach a similar band of final perplexity, the amortized optimizers are substantially more sample- and time-efficient. SABBO typically identifies a near-optimal topic number after essentially a single evaluation, and PABBO finds competitive configurations within a few evaluations, whereas GA and ES require almost the full budget to approach the same region. 
\end{abstract}

\section{Introduction}

Topic modeling refers to a class of methods that automatically discover latent thematic structure in large text collections. The core idea is that each document can be expressed as a mixture of several topics, while each topic corresponds to a probability distribution over words. Such models are widely used for document clustering, exploratory text analysis, information retrieval, and interpretability in large corpora. Among probabilistic topic models, Latent Dirichlet Allocation (LDA) remains one of the most established approaches, and the quality of its results strongly depends on the choice of the number of topics $T$.

This work addresses the optimization of the key hyperparameter $T$ (the number of topics) in the Latent Dirichlet Allocation (LDA) topic modeling approach~\cite{blei2003lda}, which is widely used for factorizing a ``document $\times$ word'' matrix as follows:
\[
[\text{Documents} \times \text{Words}] \approx [\text{Documents} \times \text{Topics}] \times [\text{Topics} \times \text{Words}].
\]
The quality of the topic model is evaluated using the standard metric of perplexity, which directly depends on the choice of the number of topics $T$ and the hyperparameters of the prior distributions $\vec{\alpha}$, $\vec{\beta}$. In this work, the hyperparameters are fixed as $\alpha=\beta=1/T$, which significantly simplifies the search procedure: the target function takes the form $f(T)$, where $f$ is the procedure for building and validating LDA for the selected $T$.

Since the analytical form and gradients of the function $f(T)$ are unavailable, the problem reduces to black-box optimization over the variable $T$. Hyperparameter tuning via black-box optimization has been widely studied in the context of machine learning, for example, using Gaussian-process-based Bayesian optimization~\cite{snoek2012practical}. Motivated by this black-box setting, we consider the following four optimization strategies and conduct a systematic comparison of their effectiveness:

\begin{itemize}
    \item \textbf{Evolution Strategy (ES):} iterative improvement of solutions is performed through mutations (random changes in the number of topics) and the selection of the best individuals from a combined population of parents and offspring, following standard practices in evolutionary computation~\cite{eiben2003intro}.
    \item \textbf{Genetic Algorithm (GA):} selection for the next generation uses tournament selection, in which the best individuals from the current generation compete with the best individuals from the previous generation. For generating new candidates, binary crossover is applied — a procedure where the binary representations of the topic count $T$ are combined, enabling offspring to inherit properties from both parents.
    \item \textbf{Preferential Amortized Black-Box Optimization (PABBO):} optimization is performed using only pairwise preference-based feedback, where the optimizer receives responses such as ``point $x$ is better than point $x'$'' rather than numerical evaluations. A neural surrogate model learns to estimate the probability that one candidate is better than another, and reinforcement learning (RL) is used to train a policy for selecting new candidates~\cite{zhang2025pabbo}. The approach allows for rapid adaptation to new tasks and effectively searches for optima when only comparative judgments are available.
    \item \textbf{Sharpness-Aware Black-Box Optimization (SABBO):} optimization is performed using sharpness-aware minimization strategy in black-box settings ~\cite{ye2025sabbo}. At each iteration, SABBO adapts the search distribution parameters by minimizing the worst-case expected objective. The algorithm applies stochastic gradient approximations using only function queries. SABBO provides theoretical guarantees of convergence and generalization, and is scalable to high-dimensional optimization tasks.

\end{itemize}

The study provides a detailed analysis of selection algorithms, mutation mechanisms, and crossover operations, and compares the effectiveness of different black-box optimization methods for tuning the number of topics in LDA with respect to the quality of topic modeling.

\section{Related Work}

\subsection{Topic Modeling}

Probabilistic topic models provide a latent, low-dimensional representation of large text corpora by modeling documents as mixtures of topics and topics as distributions over words. The most widely used baseline is Latent Dirichlet Allocation (LDA), introduced by Blei, Ng, and Jordan as a generative Bayesian model for collections of discrete data such as text corpora~\cite{blei2003lda}. In LDA, each document is represented by a multinomial distribution over topics, and each topic by a multinomial distribution over words, both regularized by Dirichlet priors.

A substantial line of work has emphasized the importance of properly choosing and tuning these Dirichlet priors. Wallach et al.\ showed that asymmetric priors over document--topic distributions can significantly improve perplexity and robustness, and that automatic hyperparameter optimization reduces the sensitivity of LDA to the number of topics while avoiding the complexity of fully nonparametric models~\cite{wallach2009rethinking}. This motivates treating LDA configuration as an explicit hyperparameter optimization problem rather than fixing priors heuristically.

Choosing the number of topics $T$ is another central challenge. A common strategy is to train models for a grid of candidate $T$ values and select the one that minimizes held-out perplexity; however, this requires many LDA fits and is sensitive to random initializations. Zhao et al. propose a heuristic based on the rate of perplexity change (RPC) as a function of $T$, providing an automated selector that tracks changes in statistical perplexity~\cite{zhao2015heuristic}. More recent work constructs composite criteria that combine perplexity with measures of topic isolation, stability across runs, and redundancy, and demonstrate improved reliability in selecting $T$ across diverse corpora~\cite{gan2021selection,neishabouri2021estimating}. These studies highlight that no single metric is universally optimal and that combining several signals often yields more robust estimates of the topic count.

Because perplexity is not always well aligned with human judgments of interpretability, alternative evaluation criteria based on topic coherence have been proposed. Newman et al. introduce automatic topic coherence scores derived from word co-occurrence statistics and show that these correlate strongly with human assessments~\cite{newman2010automatic}. Building on this idea, Mimno et al. design topic models that directly optimize a semantic coherence objective, leading to topics that are more interpretable without sacrificing statistical fit~\cite{mimno2011optimizing}. Röder et al.\ systematically compare coherence measures and propose a unifying framework that explains their behavior across datasets and models, including widely used metrics such as $C_v$~\cite{roder2015exploring}. Taken together, these works show that LDA performance is highly sensitive to hyperparameter settings and the choice of $T$, and that different evaluation metrics (perplexity, coherence, stability) capture complementary aspects of model quality. This motivates our focus on treating the selection of the number of topics as a black-box optimization problem over $T$, under a fixed budget of LDA evaluations, using perplexity-based validation as the primary objective.

\subsection{Black-Box Optimization}

Black-box optimization (BBO) studies the problem of optimizing (maximizing or minimizing) an unknown objective function when gradients and analytic structures are unavailable, and the optimizer can only access noisy function evaluations.
Classical approaches in continuous domains include evolutionary strategies and genetic algorithms, which iteratively update a population of candidate solutions using selection and stochastic variation operators~\cite{eiben2003intro}.
More advanced evolutionary strategies exist as well; for example, CMA-ES is a strong baseline for derivative-free optimization~\cite{hansen2016cmaes}.

A complementary line of work is Bayesian optimization (BO), which places a probabilistic surrogate (typically a Gaussian process or a neural surrogate) over the objective function and selects new query points by maximizing an acquisition function that trades off exploration and exploitation~\cite{snoek2012practical}.
BO has been particularly successful for expensive black-box problems such as hyperparameter tuning of machine learning models, where each function evaluation corresponds to a full training run.
However, standard BO methods are typically designed for continuous spaces and scalar-valued feedback, and often require careful engineering to handle discrete or highly structured domains.

In many applications, only relative judgments between configurations are available or more reliable than absolute scores, motivating preference-based and dueling bandit formulations of BBO.
These methods model a latent utility function and learn from pairwise comparisons $x \succ x'$ instead of numeric values $f(x)$, adapting acquisition rules and regret notions to preference feedback.
Recent work extends Bayesian optimization and bandit algorithms to this preferential setting, for both continuous and combinatorial domains~\cite{zhang2025pabbo}.

Another active direction is amortized and meta black-box optimization, where the optimizer itself is learned from a distribution of tasks so that it can rapidly adapt to new instances with limited query budgets~\cite{zhang2025pabbo,ye2025sabbo}.
Instead of handcrafting selection rules, these methods parameterize the acquisition strategy (or update rule) with a neural network and train it end-to-end using reinforcement learning or sequence modeling over optimization trajectories.
Our work follows this paradigm: we treat the choice of the number of topics $T$ as a black-box optimization problem driven by noisy perplexity evaluations, and compare hand-designed evolutionary baselines against a learned, preferential, amortized optimizer.

\section{Problem statement}

\begin{figure}[h!]
  \centering
  \includegraphics[width=0.9\textwidth]{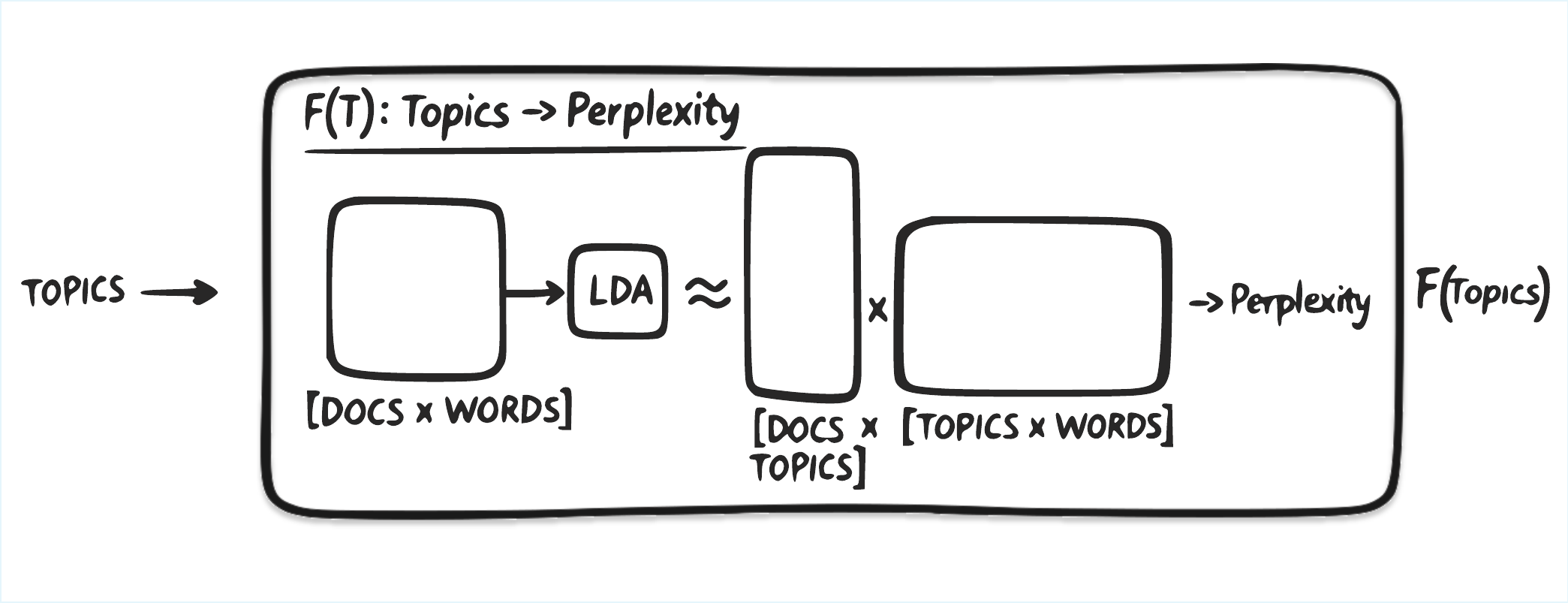}
  \caption{Problem Statement Schema}
\end{figure}

Let $W$ be a finite vocabulary indexed by $\{1, \ldots, V\}$, and $\mathcal{D}$ a corpus of $M$ documents, where each document is represented as a sequence of words from $W$.
Let $\mathcal{T}$ denote the finite set of latent topics mapped to $\lbrace 1, \dots, T\rbrace$.

The conditional independence assumption $p(w \mid d, t) = p(w \mid t)$ is adopted. Term-document probabilities do not depend on word order.  
Let $n_{dw}$ denote the number of occurrences of the word $w$ in document $d$, and define the document-term matrix  
$
N = (n_{dw}) \in (\mathbb{N} \cup \{0\})^{M \times V}.
$
Let $n_d$ denote the length of document $d$. For each document $d$, and for each word position $n \in \{1, \ldots, n_d\}$, a latent topic variable $z_{dn} \in \mathcal{T}$ is first drawn from the document-specific topic distribution $\theta_d$, and the observed word $w_{dn}$ is then drawn from the topic-specific word distribution $\phi_{z_{dn}}$.

The objective of topic modeling is to estimate the conditional word distribution
$$
p(w \mid d) = \sum_{t \in \mathcal{T}} p(w \mid t) \, p(t \mid d)
            = \sum_{t} \phi_{wt} \, \theta_{td},
$$
given the empirical counts encoded in $N$.

Latent Dirichlet Allocation (LDA) is a Bayesian generative probabilistic model for collections of discrete data, such as text corpora~\cite{blei2003lda}. All topics represented by multinomial distributions with parameter vectors $\phi_t$ share the same prior $\phi_t\sim \mathrm{Dirichlet}(\vec{\beta})$. The document-specific topic mixture parameter vector is assumed to have prior $\theta_d \sim \mathrm{Dirichlet}(\vec{\alpha})$. Topic assignments $z_{dn} \sim \mathrm{Multinomial}(\theta_d)$ and $w_{dn} \sim \mathrm{Multinomial}(\phi_{z_{dn}})$.

Given hyperparameters $\vec{\alpha}$, $\vec{\beta}$, and the number of topics $T$, for a single document $d$, the joint distribution of the topic mixture $\theta$, the topic assignments $\vec{z}$, and the observed words $\vec{w}$ is expressed as
$$
p(\theta, \vec{z}, \vec{w} \mid \vec{\alpha}, \vec{\beta})
= p(\theta \mid \vec{\alpha})
  \prod_{n=1}^{n_d} p(z_n \mid \theta) \, p(w_n \mid z_n, \vec{\beta}).
$$

Thus, the marginal likelihood of the corpus is given by
$$
p(\mathcal{D} \mid \vec{\alpha}, \vec{\beta})
= \prod_{d=1}^{M} \int p(\theta_d \mid \vec{\alpha})
   \left( \prod_{n = 1}^{n_d} \sum_{z_{dn}} p(z_{dn} \mid \theta_d) \, p(w_{dn} \mid z_{dn}, \vec{\beta}) \right)
   \mathrm{d}\theta_d.
$$
The latent variables $\theta$, $\phi$, and $z$ are typically estimated via a variational inference or expectation–maximization procedure.

To assess the quality of a topic model, the \textit{perplexity} measure is commonly employed. Using the notation introduced above, the perplexity of the model $p(w \mid d)$ on a corpus $\mathcal{D}$ is defined as
$
\mathrm{Perplexity}(\mathcal{D} )
= \exp\!\left(
    -\frac{\sum_{d=1}^{M} \sum_{w=1}^{V} n_{dw} \, \ln p(w \mid d)}
          {\sum_{d=1}^{M} \sum_{w=1}^{V} n_{dw}}
  \right).
$
Lower perplexity values correspond to models that better fit the observed data.

The present study addresses the optimization problem
$$
\mathrm{Perplexity}(\mathcal{D}) \;\to\; \min_{T \in T_{\mathrm{sanity}}},
$$
where $T_{\mathrm{sanity}} \subset \mathbb{N}$ denotes a finite, yet computationally intractable, set of admissible topic numbers for exhaustive search.  
Perplexity is evaluated for an LDA model trained on $\mathcal{D}$ with $T$ topics, under fixed hyperparameters $\alpha$ and $\beta$.

\section{Methodology}
\subsection{Genetic Algorithm}
The Genetic Algorithm (GA) maintains a population 
\[
\mathcal{P}^{(g)} = \{x^{(g)}_1, \dots, x^{(g)}_\mu\}
\]
of $\mu$ candidate solutions at generation $g$.  
Each candidate is assigned a fitness value determined by the latent objective \(f\).  
A new population is constructed via three operators:

\[
\mathcal{P}_{\text{sel}}^{(g)} = \mathrm{Select}\big(\mathcal{P}^{(g)}\big),
\qquad
x^{\text{child}} = \mathrm{Crossover}(x^{\text{par}_1}, x^{\text{par}_2}),
\qquad
x^{\text{mut}} = x^{\text{child}} + \varepsilon,
\]
    
\begin{itemize}
    \item \textbf{Selection.} A subset of parents is sampled from the current population $\mathcal{P}_{\text{sel}}^{(g)}$, using a fitness-based rule such as tournament selection or proportional selection.
    \item \textbf{Crossover.} Pairs of selected parents produce offspring via a recombination operator, which mixes coordinates of two parents to explore new regions of the domain.
    \item \textbf{Mutation.} Each offspring undergoes a small perturbation ($\varepsilon$), where $\varepsilon$ is a discrete random modification followed by clipping to the search interval.
\end{itemize}

The next population $\mathcal{P}^{(g+1)}$ is formed by replacing the previous one with all newly generated offspring (or with a mixture of elites and offspring in elitist variants).  
This evolutionary cycle continues for a fixed number of generations or until convergence.

\subsection{Evolution Strategy}

The Evolution Strategy (ES) is a family of stochastic search methods that iteratively update a small parent population by generating and selecting mutated offspring. At generation $g$ the algorithm maintains a set of $\mu$ parents $\mathcal{P}^{(g)}$ and produces $\lambda$ offspring $\mathcal{O}^{(g)}$:
\[
\mathcal{P}^{(g)} = \{x^{(g)}_1, \dots, x^{(g)}_\mu\} \subset \mathcal{X},
\qquad
\mathcal{O}^{(g)} = \{x^{(g)}_{\mu+1}, \dots, x^{(g)}_{\mu+\lambda}\}.
\]

Each offspring is sampled by applying a stochastic variation operator (mutation, optionally combined with recombination) to one or several parents:
\[
x_{\text{child}} = \mathrm{Var}\big(x^{(g)}_{p_1}, \dots, x^{(g)}_{p_k};\, \xi\big),
\]
where $\xi$ denotes injected noise and $\mathrm{Var}$ may implement, for example, additive perturbations in $\mathbb{R}^d$ or discrete modifications in a combinatorial domain. This operator controls the exploration behaviour of the strategy.

Selection then constructs the next parent population from parents and/or offspring according to their objective values:
\[
\mathcal{P}^{(g+1)} = \mathrm{Select}\big(\mathcal{P}^{(g)}, \mathcal{O}^{(g)}\big), \qquad \lvert \mathcal{P}^{(g+1)} \rvert = \mu.
\]
Classical schemes include $(\mu,\lambda)$--selection, where parents are discarded and only offspring compete, and $(\mu+\lambda)$--selection, where both parents and offspring can survive. Different ES variants further refine the variation and selection operators (e.g., recombination, step-size or covariance adaptation), but all follow this generate--evaluate--select loop.
\subsection{Preferential Amortized Black-Box Optimization}

Preferential Amortized Black-Box Optimization (PABBO)~\cite{zhang2025pabbo} tackles black-box optimization when only pairwise preferences are available instead of numeric function values. For a given task, we assume an unknown latent objective \(f : \mathcal{X} \to \mathbb{R}\) defined on \(\mathcal{X} \subset \mathbb{R}^d\). The optimizer can only query comparisons of the form
\[
x \succ x' \quad \Longleftrightarrow \quad f(x) + \varepsilon > f(x') + \varepsilon',
\]
and observes a binary label \(l \in \{0,1\}\) indicating whether \(x\) is preferred to \(x'\).

At optimization step \(t\) the algorithm has access to a duel history
\[
H_t = \{(x_{i,1}, x_{i,2}, l_i)\}_{i=1}^{t-1},
\]
where each triple encodes the compared points and the observed preference. Based on this history, PABBO learns a stochastic policy
\[
\pi_\theta(a_t \mid H_t), \qquad a_t = (x_{t,1}, x_{t,2}) \in \mathcal{X} \times \mathcal{X},
\]
which selects the next pair of points to compare. The underlying reinforcement-learning problem is defined in trajectories \((H_1, a_1, H_2, a_2, \dots, H_H)\) with rewards
\[
r_t = \max_{1 \le i \le t} \max\{f(x_{i,1}), f(x_{i,2})\},
\qquad
J(\theta) \;=\; \mathbb{E}\Bigg[\sum_{t=1}^H \gamma^{t-1} r_t\Bigg],
\]
so that the objective is to maximize the discounted cumulative reward;
i.e., to discover a point with as high a function value as possible within a fixed query budget \(H\).

To enable amortization across tasks, PABBO is trained off-line on a distribution of optimization problems and operates on three structured sets for each meta-task:
\[
    \mathcal{D}^{(c)} = \{(x^{(c)}_{i,1}, x^{(c)}_{i,2}, l^{(c)}_i)\}_{i=1}^{m_c},
    \qquad
    \mathcal{D}^{(p)} = \{(x^{(p)}_{k,1}, x^{(p)}_{k,2}, l^{(p)}_k)\}_{k=1}^{m_p},
    \qquad
    \mathcal{D}^{(q)} = \{x^{(q)}_j\}_{j=1}^{S},
\]
    
\begin{itemize}
    \item a \emph{context set} - $\mathcal{D}^{(c)}$
    which represents an optimization history and contains past duels and their preferences;
    \item a \emph{query set} - $\mathcal{D}^{(q)}$
    whose elements are combinatorially expanded into a candidate-pair set
    from which the policy \(\pi_\theta\) chooses the next comparison;
    \item a separate \emph{prediction set} - $\mathcal{D}^{(p)}$
    used only for an auxiliary preference-prediction task. At each step it is randomly split into a prediction context \(\mathcal{D}^{(\mathrm{ctx\!-\!pred})}\) and prediction targets \(\mathcal{D}^{(\mathrm{tar\!-\!pred})}\) to avoid any reward leakage from optimization to prediction.
\end{itemize}

The model follows the conditional neural process paradigm: all duels and candidate points from \(\mathcal{D}^{(c)}\), \(\mathcal{D}^{(q)}\) and \(\mathcal{D}^{(p)}\) are embedded into a shared latent space by a data embedder \(f_{\mathrm{emb}}\), processed by a transformer block \(f_{\mathrm{tfm}}\) with masked self-attention, and then decoded by two task-specific heads. The \emph{acquisition head} \(f_a\) takes the transformer outputs for candidate pairs \((x^{(q)}_i, x^{(q)}_j)\) and produces real-valued scores \(q_{i,j}\); these scores are turned into the policy via a softmax
\[
\pi_\theta\big((x^{(q)}_i, x^{(q)}_j)\mid H_t\big)
= \frac{\exp(q_{i,j})}{\sum_{(u,v)\in Q} \exp(q_{u,v})}.
\]
The \emph{prediction head} \(f_p\) maps transformer outputs for \(\mathcal{D}^{(\mathrm{tar\!-\!pred})}\) to predicted preferences on the prediction set, trained with a binary cross-entropy loss. The overall training objective combines the reinforcement-learning loss for the acquisition policy with this auxiliary prediction loss, which stabilizes training and yields a reusable acquisition strategy for new optimization tasks.

\subsection{Sharpness-Aware Black-Box Optimization (SABBO)}
Sharpness-Aware Black-Box Optimization (SABBO)~\cite{ye2025sabbo} addresses the limitations of conventional black-box optimization algorithms that directly minimize the training loss value, which often leads to poor generalization and suboptimal model quality. Inspired by the sharpness-aware minimization (SAM) principle, SABBO introduces sharpness-awareness into the black-box optimization process by explicitly accounting for the local loss landscape curvature.

Formally, for a black-box objective function \(f: \mathcal{X} \to \mathbb{R}\) defined over \(\mathcal{X} \subset \mathbb{R}^d\), SABBO reparameterizes the optimization target by its expectation over a Gaussian distribution centered at \(\mu_t\) with covariance \(\Sigma_t\):
\[
\tilde{f}(\mu_t, \Sigma_t) = \mathbb{E}_{x \sim \mathcal{N}(\mu_t, \Sigma_t)}[f(x)].
\]
At each iteration, instead of minimizing \(\tilde{f}\) directly, the algorithm optimizes the \emph{maximum} of the objective value within a small neighborhood around the current mean in distribution space, thereby encouraging solutions that are robust to sharp minima:
\[
\mu_{t+1} \;=\; \mu_t - \eta \, \widehat{\nabla}_{\mu_t} \Big(\max_{\|\delta\| \le \rho} \tilde{f}(\mu_t + \delta, \Sigma_t)\Big),
\]
where \(\eta\) is the learning rate and \(\rho\) controls the neighborhood size. The stochastic gradient \(\widehat{\nabla}_{\mu_t}\) is approximated through Monte Carlo sampling from the Gaussian distribution.

The distribution parameters are updated iteratively, capturing both the expected performance and the local geometry of the objective surface. As a result, SABBO finds solutions that generalize better to unseen data compared to traditional black-box optimization methods. Theoretically, the algorithm achieves provable convergence and generalization guarantees. Empirically, extensive experiments on black-box prompt fine-tuning tasks confirm that SABBO substantially improves generalization performance by effectively integrating sharpness-aware principles into the black-box optimization framework.

\section{Experimental Setup}

\subsection{Data}

All experiments were conducted on four text corpora commonly used for benchmarking topic models and document classification:

\begin{itemize}
    \item \textbf{20 Newsgroups}~\cite{newsgroups20_sklearn}:  
    A collection of approximately 18{,}000 posts from 20 online newsgroups, widely used as a standard testbed for text clustering and topic modeling. We use the cleaned version provided by \texttt{scikit-learn}.

    \item \textbf{AG News}~\cite{agnews_tfds}:  
    A curated subset of the AG's English news corpus, consisting of news articles categorized into four topics (World, Sports, Business, Sci/Tech). We use the \texttt{AG News Subset} version distributed through TensorFlow Datasets.

    \item \textbf{Yelp Reviews}~\cite{yelp_dataset_github}:  
    A large-scale review corpus containing user-generated restaurant and business reviews from Yelp. Following common practice, we use only the textual review fields and process them as an unsupervised topic modeling corpus.

    \item \textbf{Val\_out}:  
    A custom validation corpus constructed specifically for this study.  
    To form this dataset, we randomly sampled and merged documents from the three corpora above (20 Newsgroups, AG News, and Yelp Reviews), ensuring a balanced mixture of document lengths and topical diversity.  
    Val\_out serves as an additional out-of-distribution evaluation set for assessing the robustness and stability of topic selection across heterogeneous text sources.

\end{itemize}

Each dataset is preprocessed with lowercase normalization, stop-word removal, and token filtering, and then converted into a sparse document--term matrix using \texttt{CountVectorizer}. Vocabulary size is capped at 10{,}000 most frequent terms.

\subsection{Algorithms}
All algorithms treat evaluation of the LDA validation perplexity at a given number of topics $T$ as a black-box query. 
Throughout this section, we denote by $G$ the per-run optimization budget (number of iterations/steps).

\subsubsection{Genetic Algorithm (GA)}

Genetic Algorithm maintains a population of $\mu$ candidates, evolving them via tournament selection, binary crossover with probability $p_{\mathrm{cross}}$, mutation with probability $p_{\mathrm{mut}}$, and elitism preserving the top $N_{\mathrm{elite}}$ individuals. Crossover combines binary representations of parent $T$ values via one-point crossover at bit position $j$ (for example for 10-bit integers): offspring inherit the upper $j$ bits from one parent and lower $(10 - j)$ bits from the other, producing two children:
\[ 
    T_1^{\text{child}} = (T_1 \land \text{mask}) \lor (T_2 \land \neg\text{mask}),
    \qquad
    T_2^{\text{child}} = (T_2 \land \text{mask}) \lor (T_1 \land \neg\text{mask}),
    \qquad
    \text{mask} = (2^{j} - 1) \ll (10 - j)
\]
Mutation perturbs $T$ via a bounded random walk
\[
T^{\text{mut}} = \mathrm{clamp}(T + \varepsilon, T_{\min}, T_{\max}),
\qquad
\varepsilon \sim \mathcal{U}_{\text{discrete}}\{-\Delta T, \ldots, \Delta T\},
    \qquad
    \Delta T = 5,
\]
i.e., $\varepsilon$ is drawn from a discrete uniform distribution over integers in $[-\Delta T, \Delta T]$.

\begin{algorithm}[H]
\caption{Genetic Algorithm for optimizing $T$}
\label{alg:ga}
\begin{algorithmic}[1]
\Require population size $\mu$, number of elites $N_{\mathrm{elite}}$,
tournament size $k$, mutation rate $p_{\mathrm{mut}}$,
search interval $[T_{\min}, T_{\max}]$, number of generations $G$
\State Initialize $\mathcal{P}^{(0)} = \{T_i^{(0)}\}_{i=1}^{\mu}$ by sampling uniformly from $[T_{\min}, T_{\max}] \cap \mathbb{Z}$
\For{$g = 0$ \textbf{to} $G-1$}
    \State Evaluate fitness $f(T_i^{(g)})$ for all $T_i^{(g)} \in \mathcal{P}^{(g)}$
    \State Sort $\mathcal{P}^{(g)}$ in decreasing order of $f$
    \State $\mathcal{P}^{(g+1)} \gets$ top $N_{\mathrm{elite}}$ individuals from $\mathcal{P}^{(g)}$ \Comment{elitism}
    \State $\mathcal{O} \gets \emptyset$ \Comment{offspring pool}
    \While{$|\mathcal{O}| < \mu - N_{\mathrm{elite}}$}
        \State $p_1 \gets \textsc{TournamentSelect}(\mathcal{P}^{(g)}, k)$
        \State $p_2 \gets \textsc{TournamentSelect}(\mathcal{P}^{(g)}, k)$
        \State $(c_1, c_2) \gets \textsc{Crossover}(p_1, p_2)$
        \State $c_1 \gets \textsc{Mutate}(c_1, p_{\mathrm{mut}}, [T_{\min}, T_{\max}])$
        \State $c_2 \gets \textsc{Mutate}(c_2, p_{\mathrm{mut}}, [T_{\min}, T_{\max}])$
        \State $\mathcal{O} \gets \mathcal{O} \cup \{c_1, c_2\}$
    \EndWhile
    \State Evaluate $f$ for all $c \in \mathcal{O}$
    \State Add the best $\mu - N_{\mathrm{elite}}$ individuals from $\mathcal{O}$ to $\mathcal{P}^{(g+1)}$
\EndFor
\State \Return $\arg\max_{T \in \mathcal{P}^{(G)}} f(T)$
\end{algorithmic}
\end{algorithm}

\subsubsection{Evolution Strategy (ES).}

The Evolution Strategy keeps a parent set $\mathcal{P}^{(g)}$ of size $\mu$ and in each generation $g$ produces $\lambda$ offspring $\mathcal{O}^{(g)}$ by mutating randomly chosen parents:
\[
T_{\text{child}} = T_{\text{parent}} + \varepsilon,
\]
where $\varepsilon$ is a small zero-mean integer perturbation; the result is rounded and clipped to the search interval.
Selection uses the $(\mu + \lambda)$ scheme: the next parents are the $\mu$ best individuals from parents and offspring, focusing the search around the current best region.

\begin{algorithm}[H]
\caption{Evolution Strategy $(\mu + \lambda)$ for optimizing $T$}
\label{alg:es}
\begin{algorithmic}[1]
\Require parent population size $\mu$, offspring size $\lambda$, mutation scale $\sigma$,
search interval $[T_{\min}, T_{\max}]$, number of generations $G$, black-box objective $f(T)$
\State Initialize parent set $\mathcal{P}^{(0)} = \{T^{(0)}_j\}_{j=1}^{\mu}$ from the shared initial pool
\For{$g = 0$ \textbf{to} $G-1$}
    \State Evaluate fitness $f(T^{(g)}_j)$ for all $T^{(g)}_j \in \mathcal{P}^{(g)}$
    \State $\mathcal{O}^{(g)} \gets \emptyset$ \Comment{offspring set}
    \While{$|\mathcal{O}^{(g)}| < \lambda$}
        \State Sample parent index $j$ uniformly from $\{1, \dots, \mu\}$
        \State Sample perturbation $\varepsilon \sim \mathcal{N}(0, \sigma^2)$
        \State $T_{\text{child}} \gets \mathrm{round}\bigl(T^{(g)}_j + \varepsilon\bigr)$
        \State $T_{\text{child}} \gets \mathrm{clip}\bigl(T_{\text{child}}, T_{\min}, T_{\max}\bigr)$
        \State $\mathcal{O}^{(g)} \gets \mathcal{O}^{(g)} \cup \{T_{\text{child}}\}$
    \EndWhile
    \State Evaluate fitness $f(T)$ for all $T \in \mathcal{O}^{(g)}$
    \State $\mathcal{P}^{(g+1)} \gets$ the $\mu$ best individuals from $\mathcal{P}^{(g)} \cup \mathcal{O}^{(g)}$ according to $f$
\EndFor
\State \Return $\displaystyle \arg\max_{T \in \mathcal{P}^{(G)}} f(T)$
\end{algorithmic}
\end{algorithm}

\subsubsection{PABBO}
PABBO leverages a pre-trained Transformer model that learns from preference-based feedback.
Given optimization history $H_t$, the model predicts promising regions and selects the next $T$ via an acquisition function balancing exploration and exploitation.
The Transformer is meta-trained on synthetic functions (GP1D, Rastrigin) enabling zero-shot transfer to LDA optimization.

\begin{algorithm}[H]
\caption{PABBO-based optimization of the topic number $T$}
\label{alg:pabbo}
\begin{algorithmic}[1]
\Require pre-trained policy parameters $\theta$; exploration rate $\rho$;
query budget $G$; candidate set size $K$; search interval $[T_{\min}, T_{\max}]$; black-box
perplexity oracle $\mathrm{ppl}(T)$
\State $H_1 \gets \emptyset$ \Comment{history of evaluated topic numbers}
\For{$t = 1$ \textbf{to} $G$}
    \State Sample candidate set $C_t = \{T^{(t)}_1, \dots, T^{(t)}_K\}$ uniformly from $[T_{\min}, T_{\max}] \cap \mathbb{Z}$
    \State Internally convert $H_t$ into pairwise preferences and embed $(H_t, C_t)$ using $f_{\mathrm{emb}}$ and $f_{\mathrm{tfm}}$
    \State Obtain acquisition scores $\{q^{(t)}_k\}_{k=1}^K$ for candidates via $f_a$
    \State Define $p^{(t)} = \mathrm{softmax}(q^{(t)})$ over $C_t$
    \State Draw $b_t \sim \mathrm{Bernoulli}(\rho)$
    \If{$b_t = 1$} \Comment{exploration}
        \State Sample $T_t$ uniformly from $[T_{\min}, T_{\max}] \cap \mathbb{Z}$
    \Else \Comment{exploitation}
        \State Sample $T_t$ from $C_t$ according to $p^{(t)}$
    \EndIf
    \State Evaluate $y_t \gets \mathrm{ppl}(T_t)$
    \State Update history $H_{t+1} \gets H_t \cup \{(T_t, y_t)\}$
\EndFor
\State \Return $\displaystyle \arg \min_{(T, y) \in H_{G+1}} y$
\end{algorithmic}
\end{algorithm}
\subsubsection{SABBO}
Sharpness-Aware Black-Box Optimization (SABBO) introduces sharpness-awareness into the black-box optimization process, mitigating issues of poor generalization inherent in standard loss-minimization strategies. The methodology is deeply inspired by Sharpness-Aware Minimization (SAM), and the SABBO framework is formally described via two algorithmic variants ~\cite{ye2025sabbo} .

\vspace{1em}
\noindent
\textbf{Algorithm 1: SABBO Core Method.}
\begin{algorithm}[H]
\caption{SABBO: Sharpness-Aware Black-Box Optimization}
\label{alg:sabbo}
\begin{algorithmic}[1]
\Require Initial search mean $\mu_1$, covariance $\Sigma_1$; learning rate $\eta$; sharpness radius $\rho$; total steps $G$; black-box function $f(\cdot)$
\For{$t = 1$ \textbf{to} $G$}
    \State Sample $K$ candidate points $\{x^{(t)}_k\}_{k=1}^{K} \sim \mathcal{N}(\mu_t, \Sigma_t)$
    \For{each candidate $x^{(t)}_k$}
        \State Evaluate $f(x^{(t)}_k)$
    \EndFor
    \State Find $\tilde{x}_t = \arg\max_{\| x - \mu_t \| \le \rho} f(x)$
    \State Compute ascent direction $\widehat{\nabla}_{\mu_t} f(\tilde{x}_t)$ via stochastic approximation
    \State Update mean: $\mu_{t+1} \gets \mu_t - \eta \widehat{\nabla}_{\mu_t} f(\tilde{x}_t)$
    \State Optionally update covariance $\Sigma_{t+1}$
\EndFor
\State \Return $\mu_{G+1}$ or the best candidate found
\end{algorithmic}
\end{algorithm}

\vspace{1em}
\noindent
\textbf{Algorithm 2: SABBO (Distribution Update variant).}
\begin{algorithm}[H]
\caption{SABBO (Distribution Update variant)}
\label{alg:sabbo-dist}
\begin{algorithmic}[1]
\Require Initialize $\mu_1, \Sigma_1$; learning rate $\eta$; radius $\rho$; steps $G$
\For{$t = 1$ \textbf{to} $G$}
    \State Draw mini-batch $X_t = \{x_{t,1},\ldots x_{t,m}\} \sim \mathcal{N}(\mu_t, \Sigma_t)$
    \State For each $x_{t,i}$, generate neighbors $\mathcal{N}_\rho(x_{t,i})$
    \State For each neighbor, evaluate $f$
    \State Estimate max-neighbor: $\hat{x}_{t,i} = \arg\max_{x' \in \mathcal{N}_\rho(x_{t,i})} f(x')$
    \State Estimate gradients w.r.t.\ $\mu_t, \Sigma_t$ over $\{\hat{x}_{t,i}\}$
    \State Update $\mu_{t+1} \gets \mu_t - \eta \widehat{\nabla}_{\mu_t}$, $\Sigma_{t+1} \gets \Sigma_t - \eta \widehat{\nabla}_{\Sigma_t}$
\EndFor
\State \Return final distribution parameters $\mu_{G+1}, \Sigma_{G+1}$
\end{algorithmic}
\end{algorithm}

\vspace{1em}
Among these, we adopt the first (core) algorithm as its formulation directly matches the logical workflow of sharpness-aware black-box optimization---it searches for robust solutions by maximizing the objective in a neighborhood of the current mean, then updates the distribution parameters via a stochastic approximation. This approach is both principled and widely applicable to practical problems, whereas the second variant mainly provides a generalized scheme for distributional updates. Therefore, our further experiments and methodology focus on Algorithm 1 as best reflecting the theoretical motivation and empirical practice of SABBO.

\subsection{Setup}

LDA is trained with online variational inference. We run 10 independent trials for each dataset. For each run, we fix the random seed for LDA. All algorithms share the same initial population, sampled once from the search space of $T$. In each run, every algorithm receives a fixed budget of $G$ optimization iterations; early stopping is disabled so that all methods perform exactly $G$ optimization steps per run and operate under the same iteration budget. In PABBO we use the pre-trained Transformer policy described above.

\subsection{Computational Resources}

The research was carried out using the MSU-270 supercomputer of Lomonosov Moscow State University.

\section{Results}
\label{sec:result}

\begin{figure}[h!]
    \centering

    \begin{subfigure}{0.24\linewidth}
        \centering
        \includegraphics[width=\linewidth]{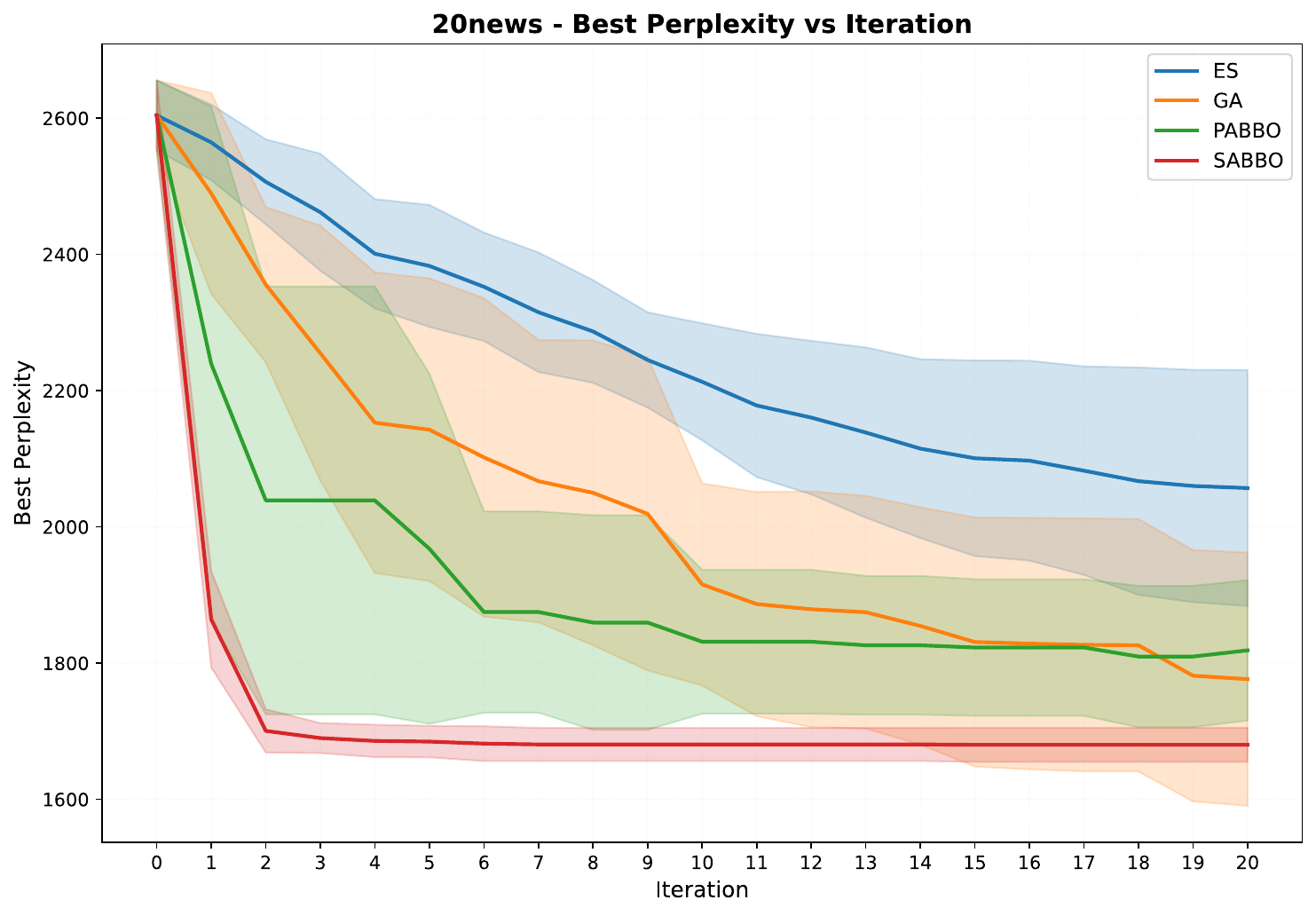}
    \end{subfigure}
    \begin{subfigure}{0.24\linewidth}
        \centering
        \includegraphics[width=\linewidth]{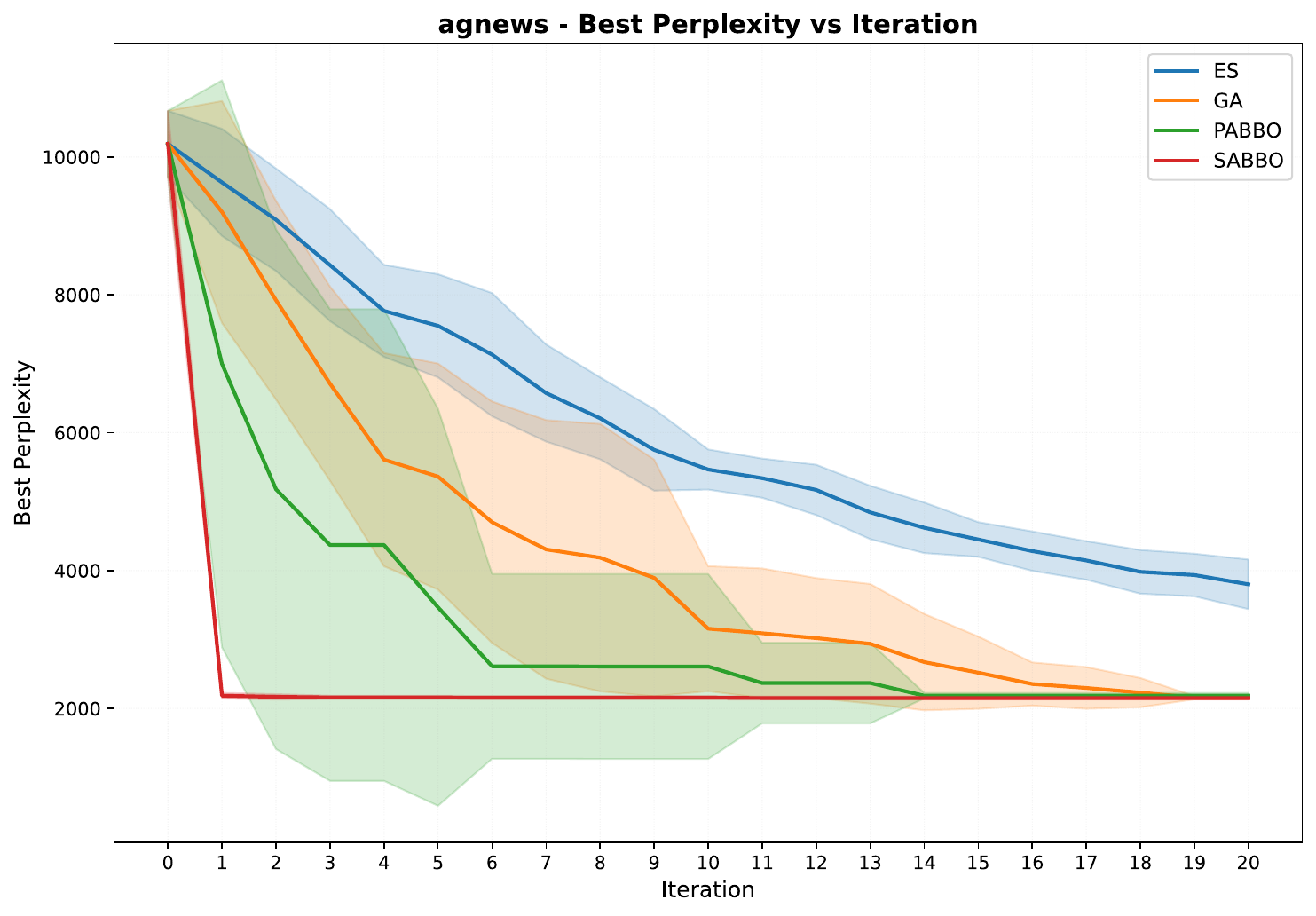}
    \end{subfigure}
    \begin{subfigure}{0.24\linewidth}
        \centering
        \includegraphics[width=\linewidth]{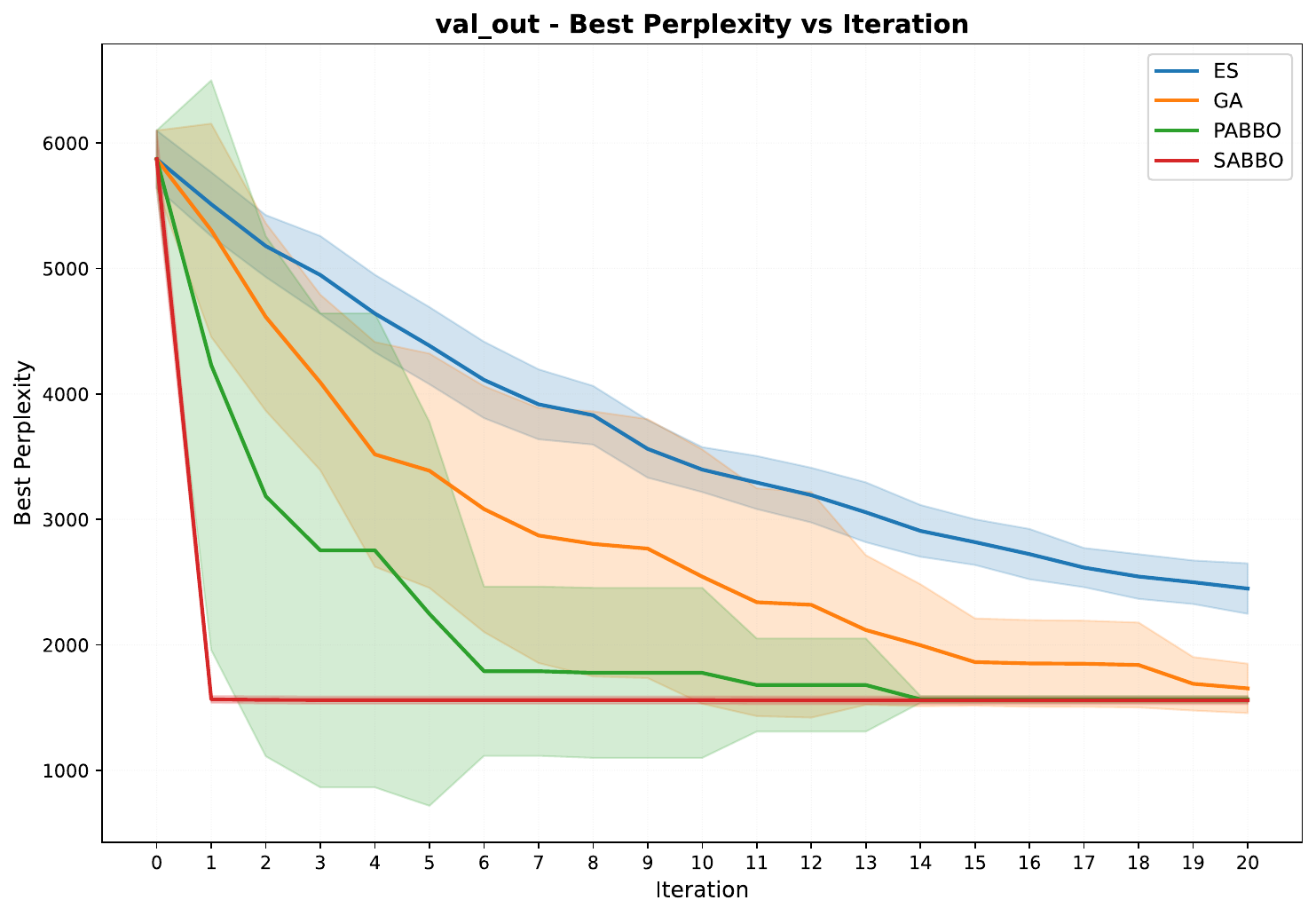}
    \end{subfigure}
    \begin{subfigure}{0.24\linewidth}
        \centering
        \includegraphics[width=\linewidth]{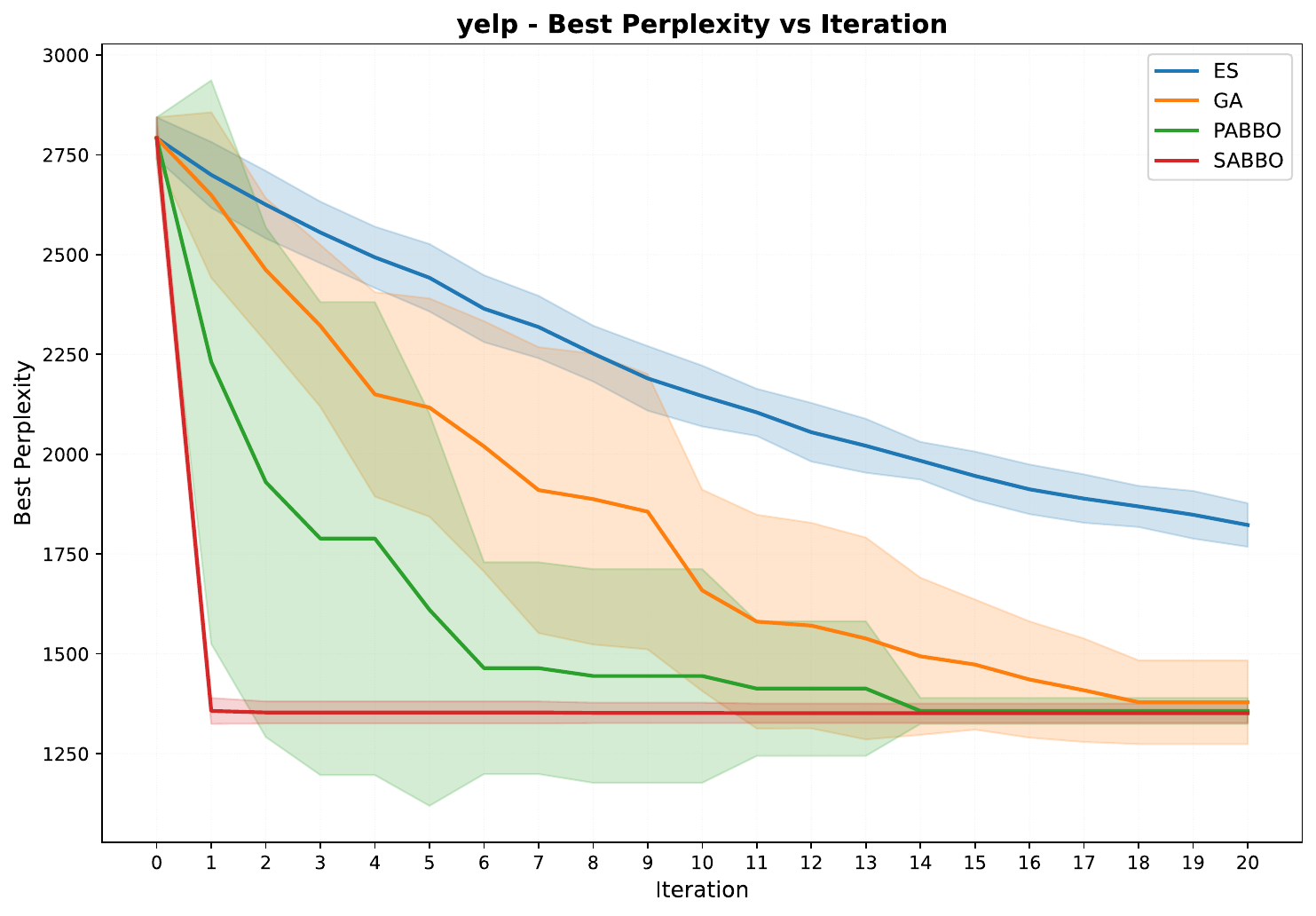}
    \end{subfigure}
    \caption{Best validation perplexity as a function of the number of LDA evaluations
    for the four corpora.}
    \label{fig:best-perplexity-iter}
\end{figure}

Figures~\ref{fig:best-perplexity-iter} and~\ref{fig:best-perplexity-time}
summarize the behaviour of the four optimizers on the four corpora.
Each curve shows, for a given method and dataset, the trajectory of the
\emph{best} validation perplexity discovered so far as the search
progresses, either as a function of the number of black-box queries
($x$--axis in Figure~\ref{fig:best-perplexity-iter}) or as a function
of cumulative wall-clock time
(Figure~\ref{fig:best-perplexity-time}).

\subsection{Convergence as a function of evaluations}

Across all datasets we observe the expected ``anytime'' behaviour:
the best perplexity decreases sharply during the first few evaluations and then exhibits diminishing returns, with only minor improvements near the end of the query budget.
After $20$ evaluations, three of the four methods (GA, PABBO, SABBO) converge to a very similar band of final perplexity, while ES remains clearly above this band but continues to move towards it.

\paragraph{GA.}
GA exhibits stable monotonic improvement but converges slowly. On all datasets the method requires nearly the full budget of 20 evaluations to reach the final perplexity band. GA consistently outperforms ES but is dominated by the learned optimizers.

\paragraph{ES.}
ES shows the weakest progress among the four methods. Its trajectories improve gradually but remain noticeably above the final perplexity levels of the other methods even after 20 evaluations. ES rarely proposes near-optimal topic numbers early in the process, leading to slower convergence.

\paragraph{PABBO.}
PABBO behaves differently.
On all datasets its trajectories contain large ``jumps'': in many runs PABBO quickly proposes a topic number $T$ that is already close to the final optimum, while in some runs the early guesses are less successful.
This is reflected in the wide confidence bands around the mean curve. Despite this variability, the average PABBO performance is consistently better than that of GA and ES and ends in the same final band as SABBO.
The stochastic policy trained by reinforcement learning therefore sometimes guesses a very good $T$ early and sometimes needs a few additional evaluations, but on average finds good configurations faster than the evolutionary baselines.

\paragraph{SABBO.}
SABBO shows the most aggressive convergence pattern.
On all corpora the perplexity curve plunges to near its final level after essentially a single evaluation, and subsequent iterations change the best value only slightly.
In other words, one SABBO query is typically enough to identify a topic number that is as good as the value obtained after running GA or ES for the full 20 iterations.
This highlights the benefit of using a sharpness-aware, amortized optimizer: most of the gain comes from the very first step, and later evaluations mainly serve to confirm the solution.

\paragraph{Early–stage performance.}
Overall, the iteration plots show that all methods except ES converge to approximately the same final perplexity level given a sufficiently large budget.
The main difference is in how quickly they get there.
SABBO almost always finds a near-optimal $T$ in the first step; PABBO often identifies a good region within a few evaluations, albeit with higher run-to-run variability; GA and ES improve steadily but slowly and only approach the same region near the end of the 20 evaluation budget.

\subsection{Convergence in wall-clock time}

\begin{figure}[h!]
    \centering

    \begin{subfigure}{0.24\linewidth}
        \centering
        \includegraphics[width=\linewidth]{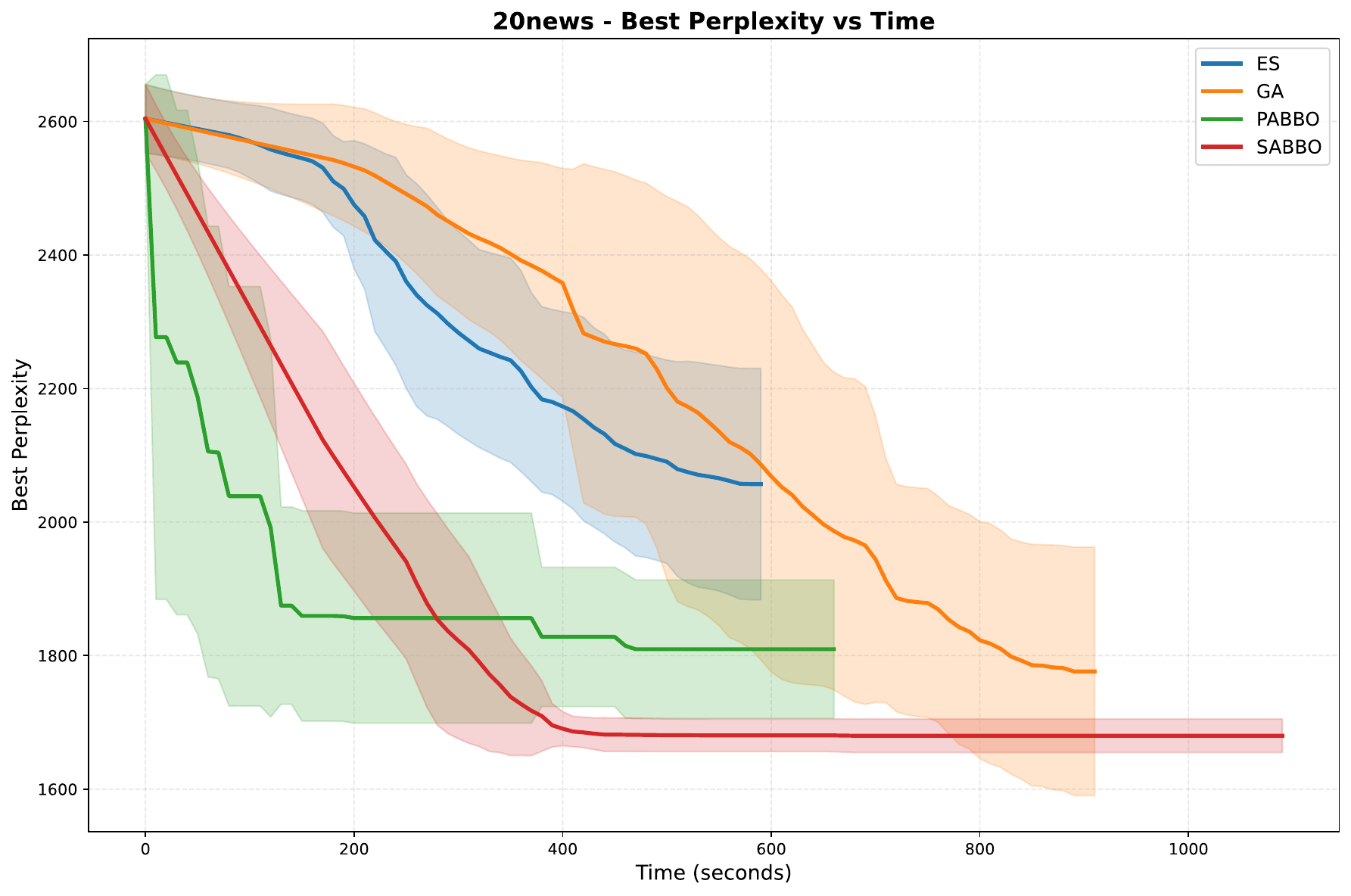}
    \end{subfigure}
    \begin{subfigure}{0.24\linewidth}
        \centering
        \includegraphics[width=\linewidth]{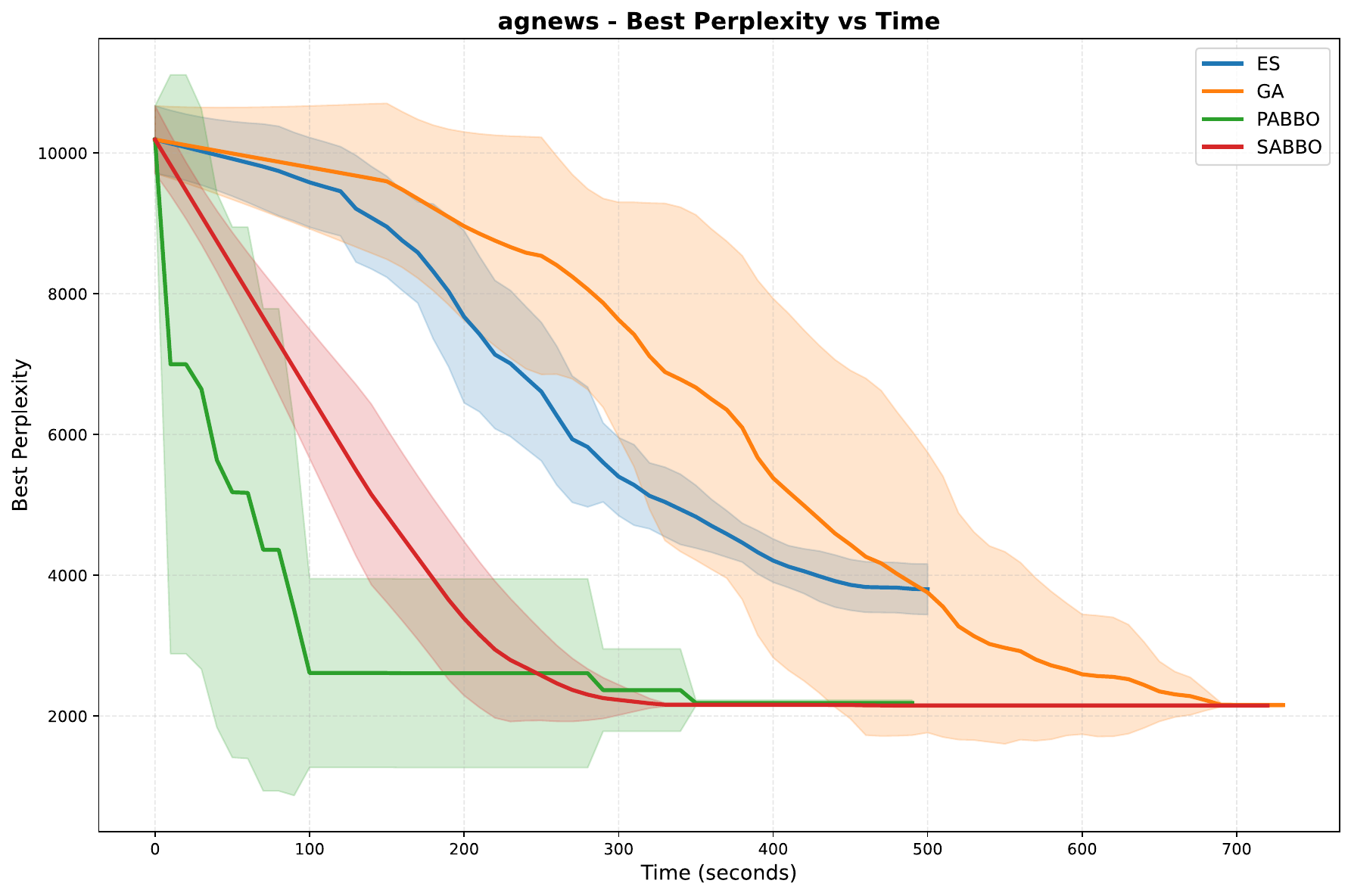}
    \end{subfigure}
    \begin{subfigure}{0.24\linewidth}
        \centering
        \includegraphics[width=\linewidth]{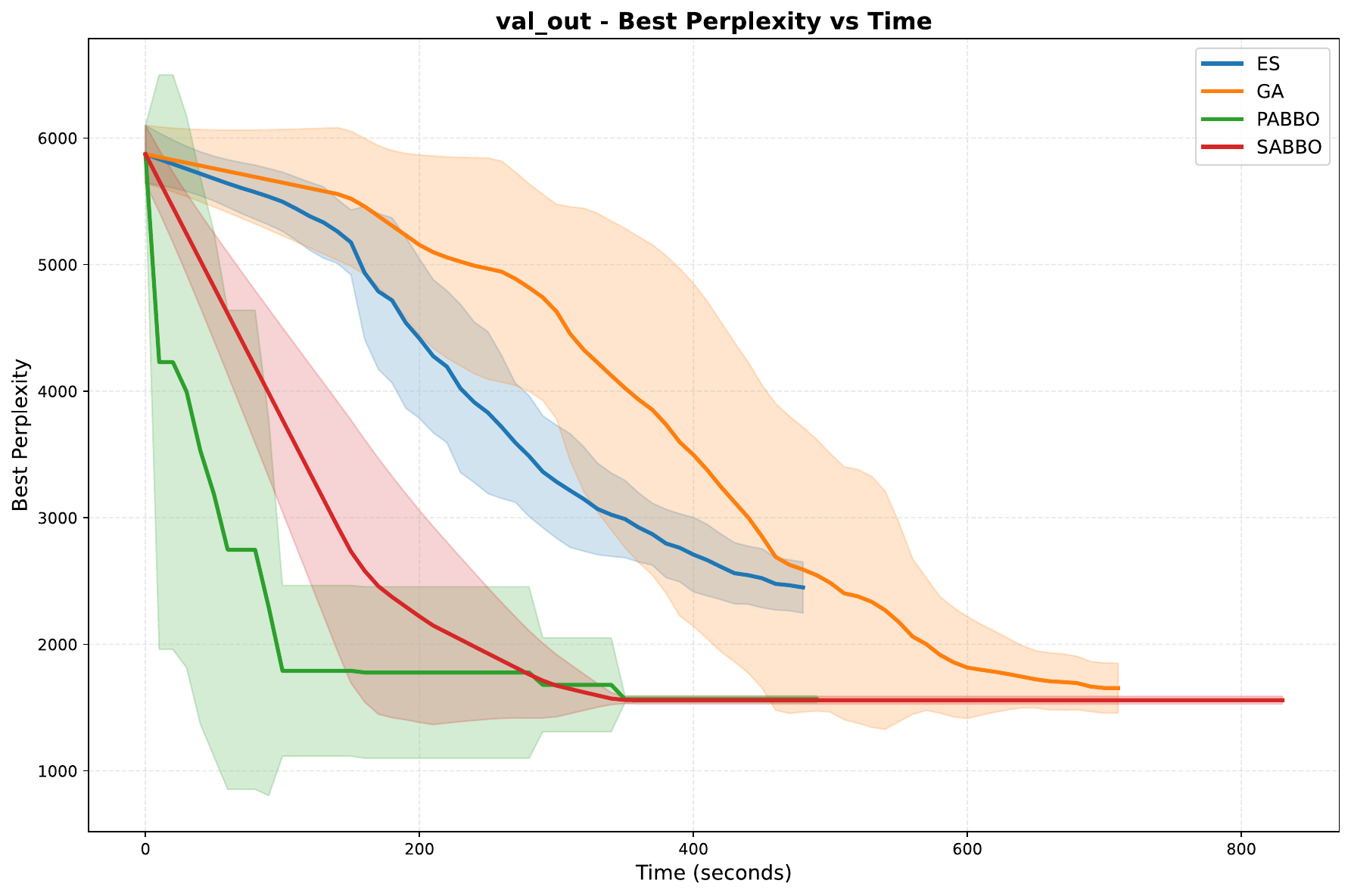}
    \end{subfigure}
    \begin{subfigure}{0.24\linewidth}
        \centering
        \includegraphics[width=\linewidth]{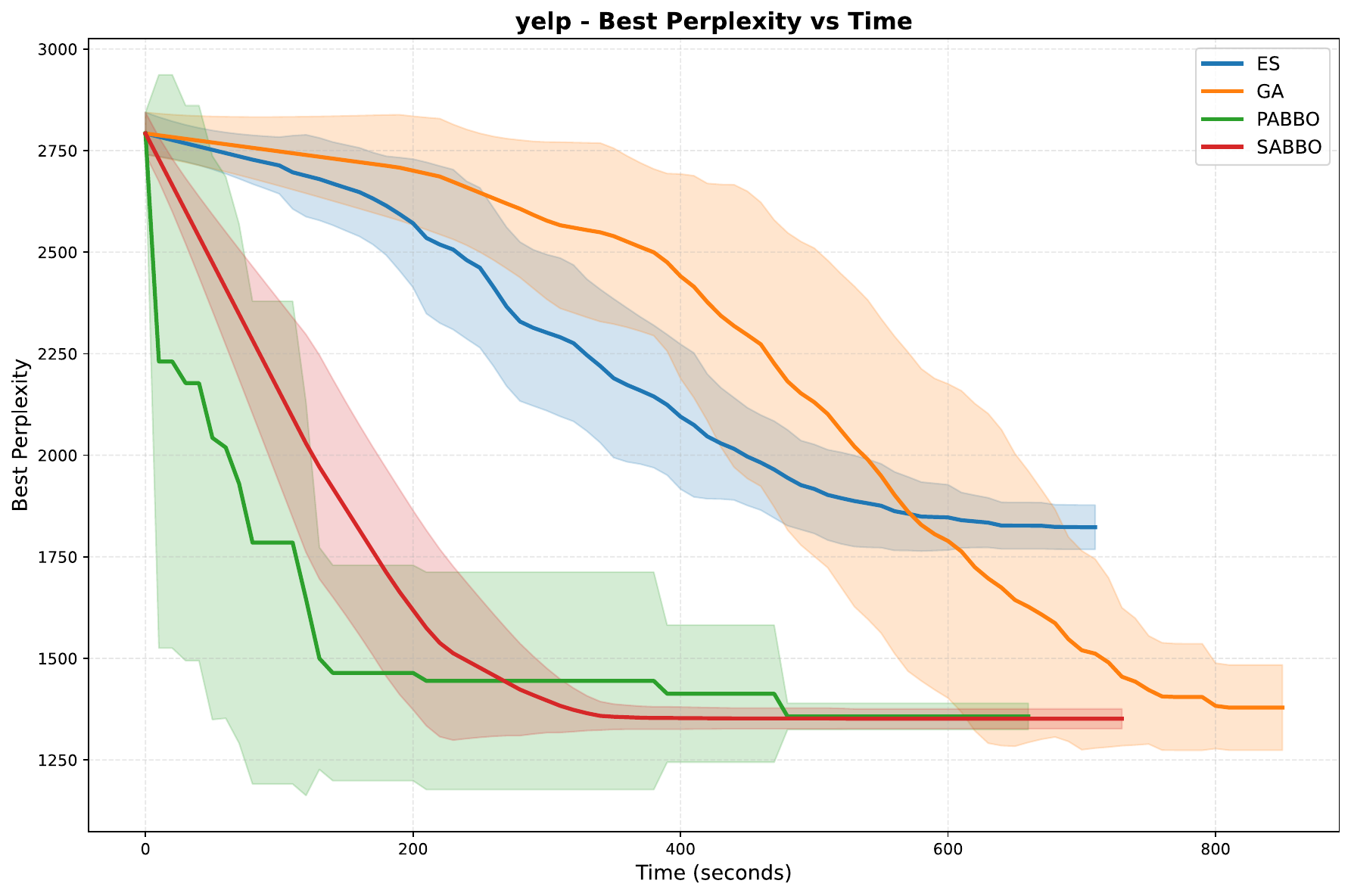}
    \end{subfigure}
    \caption{Best validation perplexity as a function of cumulative wall-clock time
    spent on LDA training and evaluation.}
    \label{fig:best-perplexity-time}
\end{figure}

Figure~\ref{fig:best-perplexity-time} reports the same trajectories as in the previous subsection, now parameterized by cumulative wall-clock time. The ranking of methods remains consistent, but the time-based view highlights important differences in computational cost.

\paragraph{GA.}
GA is consistently the slowest optimizer in terms of total runtime: completing the full budget of 20 evaluations takes approximately 30--50\% more time compared to the other methods. The method achieves competitive perplexity only after the majority of the time budget has elapsed.

\paragraph{ES.}
ES performs moderately in terms of runtime but provides weaker perplexity at all time horizons. Even when ES finishes earlier than GA, it remains significantly above the perplexity achieved by the learned optimizers at comparable time points.

\paragraph{PABBO.}
PABBO benefits from inexpensive iterations. Within the time required for a single ES or SABBO evaluation, PABBO typically performs multiple evaluations and quickly enters a good region of topic numbers. This makes it strictly better than ES in the early-stage regime under equal runtime.

\paragraph{SABBO.}
SABBO has the most expensive first evaluation, but this step is highly informative: on all datasets, the first SABBO point yields a perplexity significantly lower than the initial points of GA and ES, and often within 10--20\% of the final optimum. Subsequent evaluations refine an already competitive solution.

\subsection{Final Performance Summary}

Table~\ref{tab:final-perplexity} reports the final best perplexity (mean $\pm$ standard deviation across runs) achieved by each method on all datasets after the full evaluation budget of $20$ queries. These values correspond to the endpoints of the curves shown earlier.

\begin{table}[h!]
\centering
\begin{tabular}{lcccc}
\hline
\textbf{Dataset} & \textbf{GA} & \textbf{ES} & \textbf{PABBO} & \textbf{SABBO} \\
\hline
20NEWS   & $1776.35 \pm 185.94$ & $2056.91 \pm 173.29$ & $1809.55 \pm 103.79$ & \textbf{1679.96 $\pm$ 25.09} \\
AGNEWS   & $2154.98 \pm 23.82$  & $3800.07 \pm 359.69$ & $2184.55 \pm 40.17$  & \textbf{2150.56 $\pm$ 22.05} \\
VAL\_OUT & $1653.49 \pm 196.91$ & $2448.87 \pm 201.28$ & $1565.99 \pm 27.31$  & \textbf{1557.91 $\pm$ 30.50} \\
YELP     & $1378.81 \pm 104.88$ & $1822.72 \pm 54.55$  & $1356.98 \pm 32.48$  & \textbf{1351.21 $\pm$ 24.32} \\
\hline
\end{tabular}
\caption{Final best validation perplexity after 20 evaluations.}
\label{tab:final-perplexity}
\end{table}

\paragraph{Overall conclusions.}
\begin{itemize}
    \item SABBO achieves the best final perplexity on all four datasets, with the smallest variance.
    \item PABBO is the second-best method overall, usually close to SABBO and consistently outperforming GA and ES.
    \item GA performs moderately: better than ES and often competitive with PABBO on some datasets, but slower and less stable.
    \item ES performs the worst across all datasets, confirming that simple evolutionary strategies are not well-suited for this optimization task.
\end{itemize}

Taken together, the evaluation-based, time-based, and final perplexity analyses show that amortized black-box optimizers (PABBO and SABBO) provide clear advantages over classical evolutionary baselines in both efficiency and final model quality.

\section{Future discussion}

In the course of our experiments, we obtain a dataset consisting of multiple corpora together with their optimal numbers of topics empirically determined $T^\ast$.
This dataset provides a foundation for formulating the estimation of $T^\ast$ as a supervised learning problem.
Depending on how the target space is defined, this task can be approached either as regression, treating $T^\ast$ as an integer-valued quantity over a broad range, or as a classification over a restricted set of admissible topic numbers.
Developing such predictive models would shift the main computational burden from repeated LDA training on individual corpora to an offline model-selection stage across many corpora, enabling fast and efficient estimation of suitable topic numbers for new datasets.

Another promising direction for reducing computational overhead in topic model configuration is to cast the selection of the topic number as a reinforcement learning (RL) problem.
A corpus (or its feature representation) can be treated as the state, the choice of $T$ as the action, and a reward function can evaluate the resulting LDA model.
Unlike supervised approaches, an RL agent does not rely on precomputed optimal topic numbers—which may be costly to obtain or inherently noisy—but instead learns a policy for choosing $T$ directly from reward-based feedback.
The central challenge lies in defining a reward signal that reliably reflects the quality of the topic model, given that common evaluation measures can be unstable or only partially correlated with human judgments.
This line of research could lead to practical methods for automated topic model selection.

\section*{Acknowledgments}

This work was supported by the Ministry of Economic Development of the Russian Federation
in accordance with the subsidy agreement (agreement identifier 000000C313925P4H0002;
grant No 139-15-2025-012).

\bibliographystyle{plain}
\bibliography{references}

@article{blei2003lda,
  title   = {Latent Dirichlet Allocation},
  author  = {Blei, David M. and Ng, Andrew Y. and Jordan, Michael I.},
  journal = {Journal of Machine Learning Research},
  volume  = {3},
  pages   = {993--1022},
  year    = {2003},
  url     = {https://www.jmlr.org/papers/volume3/blei03a/blei03a.pdf}
}

@inproceedings{wallach2009rethinking,
  title     = {Rethinking {LDA}: Why Priors Matter},
  author    = {Wallach, Hanna M. and Mimno, David and McCallum, Andrew},
  booktitle = {Advances in Neural Information Processing Systems},
  volume    = {22},
  pages     = {1973--1981},
  year      = {2009},
  url       = {https://mimno.infosci.cornell.edu/papers/NIPS2009_0929.pdf}
}

@article{zhao2015heuristic,
  title   = {A Heuristic Approach to Determine an Appropriate Number of Topics in Topic Modeling},
  author  = {Zhao, Weizhong and Chen, James J. and Perkins, Roger and Liu, Zhichao and Ge, Weigong and Ding, Yijun and Zou, Wen},
  journal = {BMC Bioinformatics},
  volume  = {16},
  number  = {S13},
  pages   = {S8},
  year    = {2015},
  doi     = {10.1186/1471-2105-16-S13-S8},
  url     = {https://bmcbioinformatics.biomedcentral.com/articles/10.1186/1471-2105-16-S13-S8}
}

@article{gan2021selection,
  title   = {Selection of the Optimal Number of Topics for {LDA} Topic Model---Taking Patent Policy Analysis as an Example},
  author  = {Gan, Jingxian and Qi, Yong},
  journal = {Entropy},
  volume  = {23},
  number  = {10},
  pages   = {1301},
  year    = {2021},
  doi     = {10.3390/e23101301},
  url     = {https://www.mdpi.com/1099-4300/23/10/1301}
}

@inproceedings{neishabouri2021estimating,
  title     = {Estimating the Number of Latent Topics Through a Combination of Methods},
  author    = {Neishabouri, Asana and Desmarais, Michel C.},
  booktitle = {Proceedings of the 25th International Conference on Knowledge-Based and Intelligent Information and Engineering Systems (KES 2021)},
  series    = {Procedia Computer Science},
  volume    = {192},
  pages     = {1190--1197},
  year      = {2021},
  doi       = {10.1016/j.procs.2021.08.122},
  url       = {https://www.sciencedirect.com/science/article/pii/S1877050921016112}
}

@inproceedings{newman2010automatic,
  title     = {Automatic Evaluation of Topic Coherence},
  author    = {Newman, David and Lau, Jey Han and Grieser, Karl and Baldwin, Timothy},
  booktitle = {Human Language Technologies: The 2010 Annual Conference of the North American Chapter of the Association for Computational Linguistics},
  pages     = {100--108},
  year      = {2010},
  address   = {Los Angeles, California},
  publisher = {Association for Computational Linguistics},
  url       = {https://aclanthology.org/N10-1012/}
}

@inproceedings{mimno2011optimizing,
  title     = {Optimizing Semantic Coherence in Topic Models},
  author    = {Mimno, David and Wallach, Hanna and Talley, Edmund and Leenders, Miriam and McCallum, Andrew},
  booktitle = {Proceedings of the 2011 Conference on Empirical Methods in Natural Language Processing},
  pages     = {262--272},
  year      = {2011},
  address   = {Edinburgh, Scotland, UK},
  publisher = {Association for Computational Linguistics},
  url       = {http://www.cs.princeton.edu/~mimno/papers/mimno-semantic-emnlp.pdf}
}

@inproceedings{roder2015exploring,
  title     = {Exploring the Space of Topic Coherence Measures},
  author    = {R{\"o}der, Michael and Both, Andreas and Hinneburg, Alexander},
  booktitle = {Proceedings of the Eighth ACM International Conference on Web Search and Data Mining},
  pages     = {399--408},
  year      = {2015},
  address   = {Shanghai, China},
  publisher = {ACM},
  doi       = {10.1145/2684822.2685324},
  url       = {https://dl.acm.org/doi/10.1145/2684822.2685324}
}

@book{eiben2003intro,
  title     = {Introduction to Evolutionary Computing},
  author    = {Eiben, Agoston E. and Smith, James E.},
  year      = {2003},
  publisher = {Springer},
  address   = {Berlin, Heidelberg},
  isbn      = {3-540-40184-9}
}

@inproceedings{snoek2012practical,
  title     = {Practical Bayesian Optimization of Machine Learning Algorithms},
  author    = {Snoek, Jasper and Larochelle, Hugo and Adams, Ryan P.},
  booktitle = {Advances in Neural Information Processing Systems},
  volume    = {25},
  pages     = {2960--2968},
  year      = {2012},
  url       = {https://papers.nips.cc/paper/4522-practical-bayesian-optimization-of-machine-learning-algorithms}
}

@inproceedings{zhang2025pabbo,
  title     = {{PABBO}: Preferential Amortized Black-Box Optimization},
  author    = {Zhang, Xinyu and Huang, Daolang and Kaski, Samuel and Martinelli, Julien},
  booktitle = {Proceedings of the International Conference on Learning Representations},
  year      = {2025},
  note      = {ICLR 2025, Spotlight},
  url       = {https://arxiv.org/abs/2503.00924}
}

@inproceedings{ye2025sabbo,
  title     = {{SABBO}: Sharpness-Aware Black-Box Optimization},
  author    = {Ye, Feiyang and Lyu, Yueming and Wang, Xuehao and Sugiyama, Masashi and Zhang, Yu and Tsang, Ivor W.},
  booktitle = {Proceedings of the International Conference on Learning Representations},
  year      = {2025},
  note      = {ICLR 2025, Spotlight},
  url       = {https://openreview.net/pdf?id=h7EwIfjxgn}
}

@article{hansen2016cmaes,
  title   = {The {CMA} Evolution Strategy: A Tutorial},
  author  = {Hansen, Nikolaus},
  journal = {arXiv preprint arXiv:1604.00772},
  year    = {2016},
  doi     = {10.48550/arXiv.1604.00772},
  url     = {https://arxiv.org/abs/1604.00772}
}

@article{sklearn,
  title={Scikit-learn: Machine Learning in {P}ython},
  author={Pedregosa, F. and Varoquaux, G. and Gramfort, A. and Michel, V.
          and Thirion, B. and Grisel, O. and Blondel, M. and Prettenhofer, P.
          and Weiss, R. and Dubourg, V. and Vanderplas, J. and Passos, A. and
          Cournapeau, D. and Brucher, M. and Perrot, M. and Duchesnay, E.},
  journal={Journal of Machine Learning Research},
  volume={12},
  pages={2825--2830},
  year={2011}
}

@article{hoffman2010online,
  title={Online learning for latent dirichlet allocation},
  author={Hoffman, Matthew and Bach, Francis R and Blei, David M},
  journal={Advances in neural information processing systems},
  volume={23},
  year={2010}
}

@article{deap,
  author    = {F\'elix-Antoine Fortin and Fran\c{c}ois-Michel {De Rainville} and Marc-Andr\'e Gardner and Marc Parizeau and Christian Gagn\'e},
  title     = {{DEAP}: Evolutionary Algorithms Made Easy},
  journal   = {Journal of Machine Learning Research},
  volume    = {13},
  pages     = {2171--2175},
  year      = {2012}
}

@misc{zhang2025pabbo-code,
  author = {Zhang, Xinyu and Huang, Daolang and Kaski, Samuel and Martinelli, Julien},
  title = {{PABBO}: {P}referential {A}mortized {B}lack-{B}ox {O}ptimization - Official Implementation},
  year = {2025},
  howpublished = {\url{https://github.com/xinyuzhang99/PABBO}},
  note = {Accessed: 2024-11-21}
}

@misc{agnews_tfds,
  title        = {AG News Subset Dataset},
  author       = {Zhang, Xiang and Zhao, Junbo and LeCun, Yann},
  howpublished = {\url{https://www.tensorflow.org/datasets/catalog/ag_news_subset}},
  note         = {Accessed: 2025-12-04},
  year         = {2015}
}

@misc{newsgroups20_sklearn,
  title        = {20 Newsgroups Dataset},
  author       = {Lang, Ken},
  howpublished = {\url{https://scikit-learn.org/stable/modules/generated/sklearn.datasets.fetch_20newsgroups.html}},
  note         = {Accessed: 2025-12-04},
  year         = {1995}
}

@misc{yelp_dataset_github,
  title        = {Yelp Review Dataset},
  author       = {Yelp Inc.},
  howpublished = {\url{https://github.com/kevintee/Yelp-Dataset}},
  note         = {Accessed: 2025-12-04},
  year         = {2020}
}

\appendix

\section{Implementation}

\subsection{LDA Implementation}

We use scikit-learn's \texttt{LatentDirichletAllocation} implementation~\cite{sklearn}, which employs online variational Bayes inference~\cite{hoffman2010online}.

\subsection{Optimizer Implementations}

\paragraph{Genetic Algorithm (GA).}
Implemented using the DEAP framework (v1.4.1)~\cite{deap}. Binary crossover operates at the bit level: each integer $T$ is converted to binary representation, a crossover point is selected, and bit strings are exchanged between parents. Integer mutation adds discrete noise $\varepsilon \sim \mathcal{U}_{\text{discrete}}\{-5, \ldots, 5\}$ with clamping to $[2, 1000]$.

\paragraph{Evolution Strategy (ES).}
Also implemented using DEAP for population management. We use $(\mu + \lambda)$ selection where $\mu=5$ parents and $\lambda=10$ offspring compete, with the best $\mu$ individuals advancing. Gaussian mutation applies $T_{\text{offspring}} = T_{\text{parent}} + \mathcal{N}(0, 5^2)$ followed by rounding and clipping.

\paragraph{PABBO.}
Based on the official implementation from~\cite{zhang2025pabbo-code}. The Transformer model is implemented in PyTorch 2.3.0 with the following architecture:
\begin{itemize}
    \item Data embedder: 2-layer MLP projecting $(T, \text{perplexity})$ to 32 dimensions
    \item Transformer encoder: 3 layers, 2 attention heads, 64-dim feedforward
    \item Acquisition head: linear layer producing candidate scores
    \item Prediction head: binary classifier for preference learning
\end{itemize}

The model is trained from scratch on 1D synthetic functions (Rastrigin, GP with RBF kernel) for 10,000 episodes using Adam optimizer (learning rate $3 \times 10^{-4}$) and policy gradient with discount factor $\gamma=0.99$. The trained checkpoint is then loaded and applied to LDA optimization via a custom wrapper that maintains optimization history and queries the PABBO policy at each iteration.

\paragraph{SABBO.}
Implemented based on the official algorithm specification from~\cite{ye2025sabbo}. The method reparameterizes the objective through a Gaussian search distribution \(p_{\theta}(x)=\mathcal{N}(x|\mu,\Sigma)\) with diagonal covariance. Both full-batch and mini-batch query modes are supported. 

At each iteration, \(K=10\) samples are drawn from the current distribution to estimate the stochastic gradients \(\nabla_\mu J(\theta)\) and \(\nabla_\Sigma J(\theta)\) using Monte Carlo approximation. Perturbations \(\delta_\mu,\delta_\Sigma\) are then computed according to Eqs.~(18–21) in the paper with neighborhood size \(\rho=0.05\) and step size schedule \(\beta_t=1/t\) (differs from the article we rely on because we want all methods to predict $T$). 

We use a monotonic transformation of the queried objective values to stabilize training as suggested in Appendix~F of~\cite{ye2025sabbo}. The covariance update \(\Sigma^{-1}_{t+1} = \Sigma^{-1}_t + 2\beta_t G_t\) and mean update \(\mu_{t+1} = \mu_t - \beta_t \Sigma_t g_t\) are applied sequentially, with numerical safeguards enforcing \(\Sigma_{ii}\ge10^{-4}\).

\subsection{Data Preprocessing}

All text corpora are preprocessed using scikit-learn's \texttt{CountVectorizer} with lowercase conversion, English stop word removal, vocabulary limited to 10,000 most frequent terms, and filtering terms with document frequency below 5 or above 90\%. The resulting document-term matrices are stored in compressed sparse row (CSR) format.

\end{document}